\documentclass[letterpaper]{article}
\usepackage[preprint]{aaai2027} 
\usepackage[hyphens]{url} 
\usepackage{graphicx}
\usepackage{natbib}
\usepackage{caption}
\usepackage{algorithm}
\usepackage{algorithmic}
\usepackage{amsmath}

\usepackage{newfloat}
\usepackage{listings}
\DeclareCaptionStyle{ruled}{labelfont=normalfont,labelsep=colon,strut=off}
\floatstyle{ruled}
\newfloat{listing}{tb}{lst}{}
\floatname{listing}{Listing}

\usepackage{booktabs}

\usepackage{amssymb}
\usepackage{multirow}

\newsavebox{\UniVOEauthorbox}
\newcommand{\UniVOEauthorline}[1]{%
  \sbox{\UniVOEauthorbox}{#1}%
  \ifdim\wd\UniVOEauthorbox>\textwidth
    \resizebox{\textwidth}{!}{\usebox{\UniVOEauthorbox}}%
  \else
    \usebox{\UniVOEauthorbox}%
  \fi
}
\title{UniVOE: A Removal-Driven Framework for Interaction-Aware Universal Video Object Editing}
\author{
    \mdseries
    Shuzhou Yang\textsuperscript{1}\footnotemark[2] \quad
    Lanqing Guo\textsuperscript{2}\footnotemark[1] \quad
    Lingen Li\textsuperscript{3} \quad
    Weiqi Li\textsuperscript{1}\\[2pt]
    Zhao Zhong\textsuperscript{2} \quad
    Liefeng Bo\textsuperscript{2} \quad
    Jian Zhang\textsuperscript{1}\footnotemark[1]
}

\affiliations{
    \bfseries
    \mbox{%
        \textsuperscript{1}SECE, Peking University\qquad
        \textsuperscript{2}Tencent Hunyuan\qquad
        \textsuperscript{3}The Chinese University of Hong Kong%
    }
}

\begin{document}

\makeatletter
\let\@oldmaketitle\@maketitle
\renewcommand{\@maketitle}{%
  \@oldmaketitle
  \vspace{-3.2em}
  \begin{center}

    \mbox{%
      \normalfont
      \pdfstartlink attr{/Border [0 0 0]}
        user{%
          /Subtype /Link
          /A << /S /URI /URI (https://univoe.github.io/) >>
        }%
        \texttt{\textcolor[RGB]{70,150,210}{\texttt{https://univoe.github.io/}}%
}%
      \pdfendlink
    }\par
    \vspace{0.5em}

    \begin{minipage}{0.87\linewidth}
      \centering
      \includegraphics[width=\textwidth]{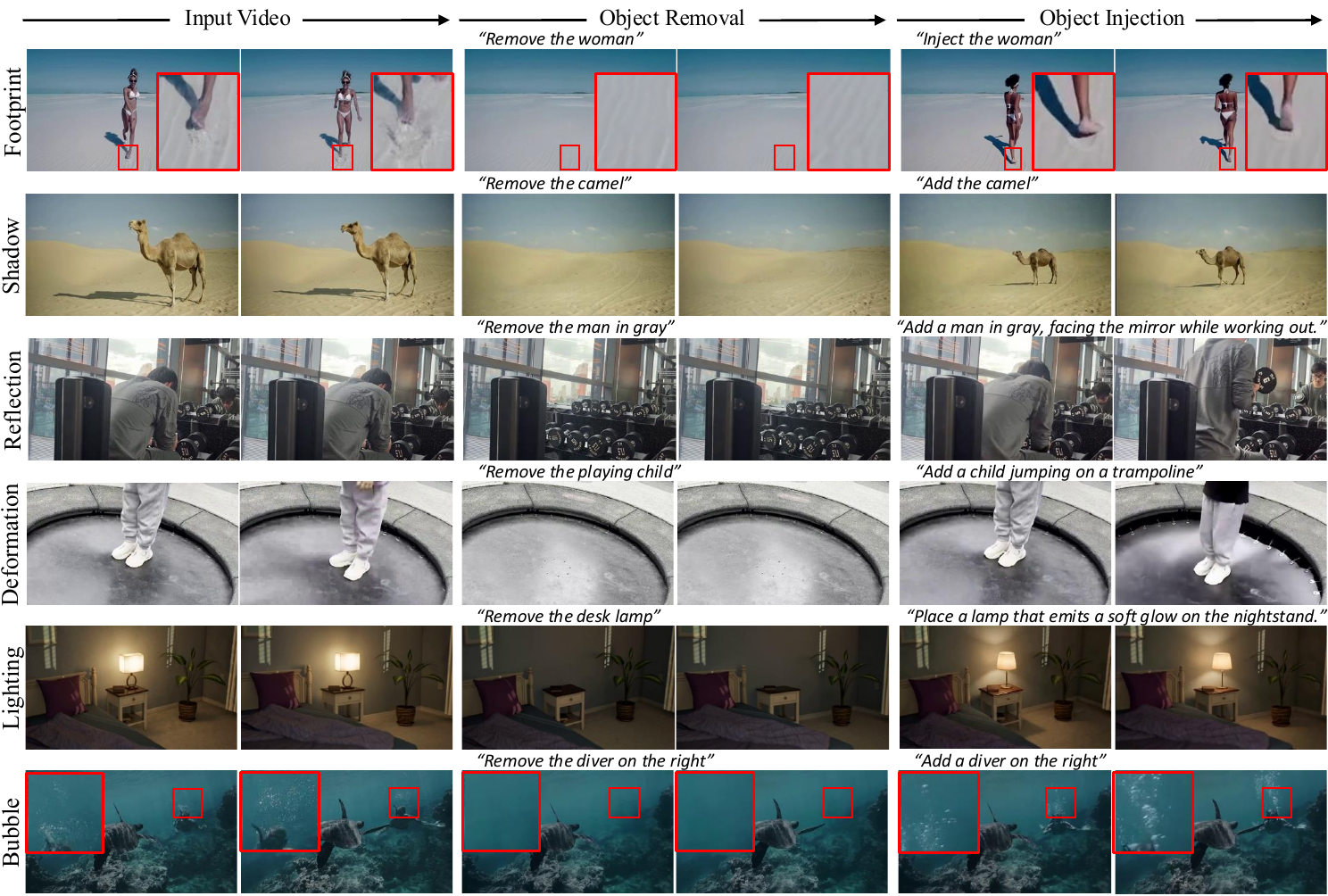}
      \vspace{-2.em}
      \captionof{figure}{We propose \textbf{UniVOE}, a video object editing pipeline that enables object-centered removal and injection, together with predictions of changes in object-related effects. Such as footprint, shadow, reflection, and lighting. Users can easily remove or insert target objects using prompts, as shown above each editing result.}
      \label{fig:teaser}
      \vspace{0.7em}
    \end{minipage}
  \end{center}
}
\makeatother

\maketitle
\begingroup
\renewcommand{\thefootnote}{\fnsymbol{footnote}}
\footnotetext[1]{Corresponding authors.}
\footnotetext[2]{Work done during an internship at Hunyuan, Tencent.}
\endgroup
\pagestyle{plain}
\thispagestyle{plain}

\begin{abstract}
Video object removal and injection are reciprocal editing operations between object-present and object-absent video states, yet they differ substantially in training difficulty. Removal can be effectively bootstrapped from video inpainting and further adapted to erase object-related interaction effects. Injection, however, requires large-scale paired supervision to model object appearance, dynamics, and scene-dependent effects, while perfectly aligned real-world pairs are difficult to collect and synthetic data lacks sufficient realism and diversity. To address this challenge, we present \textbf{UniVOE}, a removal-driven framework for training both video object removal and injection models. We first train an effect-aware remover by transferring video inpainting priors and finetuning on object-present/object-absent pairs. The trained remover converts real object-present videos into object-absent counterparts, creating large-scale paired supervision for injection. An injection model is trained on these pairs in the reverse direction to recover objects together with their dynamics and interaction effects. Moreover, since pixel-level reconstruction alone cannot adequately capture perceptual realism, we introduce a reward-based post-training method that aligns injection results with human preferences using a proprietary reward model. Experiments against strong open-source and commercial systems demonstrate that UniVOE achieves state-of-the-art performance on both object removal and injection, particularly in challenging interaction-effect cases.
\end{abstract}

\section{Introduction}
\label{sec:intro}

Video object removal and injection are critical in visual content creation, with broad applications in film post-production, advertising, and virtual production. Unlike image editing, video editing must account for not only the target object itself, but also its dynamic imprint on surrounding scene. A moving object may cast shadows, produce reflections, and contact surfaces. Editing only the object but leaving these interaction effects untreated during removal or synthesizing them incorrectly in injection will produce perceptible artifacts. A practical tool should jointly edit the object and its effects.

Recent diffusion-based video editing methods made significant progress in fidelity and temporal consistency~\cite{nvedit,diffueraser,propainter,objectwiper,4dvd}. Some works further recognize the close relationship between object removal and insertion~\cite{effecterase}: effect-aware removers erase shadows and reflections together with the target object, while reciprocal formulations use insertion as an auxiliary inverse task to improve removal. General video inpainting systems also support adding or removing objects~\cite{videopainter}. Nevertheless, realistic object injection remains difficult. It must hallucinate absent foreground, infer plausible object motion, and predict interactive effects such as shadows and reflections. The main bottleneck is supervision. Ideally, both directions should be trained with paired videos of the same scene in object present and absent states under identical lighting, and background dynamics. Such pairs are hard to capture in the wild, while synthetic rendering, despite perfect alignment, lacks diversity, photorealism, and dynamics of real footage.

Our key insight is that object removal and injection share the same paired data structure, but are asymmetric in learning difficulty. Let $V_{\text{obj}}$ and $V_{\text{abs}}$ denote the object present and absent states of the same video. The two tasks can be viewed as conditional generation in opposite directions:
$
\pi_{\text{rem}}(V_{\text{abs}}\mid V_{\text{obj}},\,c)
\Longleftrightarrow
\pi_{\text{inj}}(V_{\text{obj}}\mid V_{\text{abs}},\,c)
$,
where $c$ denotes the editing condition, such as text prompt in Fig.~\ref{fig:teaser}. The same pair can supervise both directions: removal learns to eliminate the object and its effects, while injection learns to predict them. However, removal can be scaled from video inpainting, which provides abundant supervision for background completion, temporal structure, and scene consistency. With limited paired $V_{\text{obj}}$-$V_{\text{abs}}$ data, this prior can be adapted to complete object removal and effect erasing. Injection, in contrast, requires cleaner supervision for object appearance, motion, and interaction. This reciprocal-but-asymmetric structure motivates removal-driven strategy: strengthen easier removal direction, and use it to derive supervision for injection.

\begin{figure}[t]
  \centering
    \includegraphics[width=1.\linewidth]{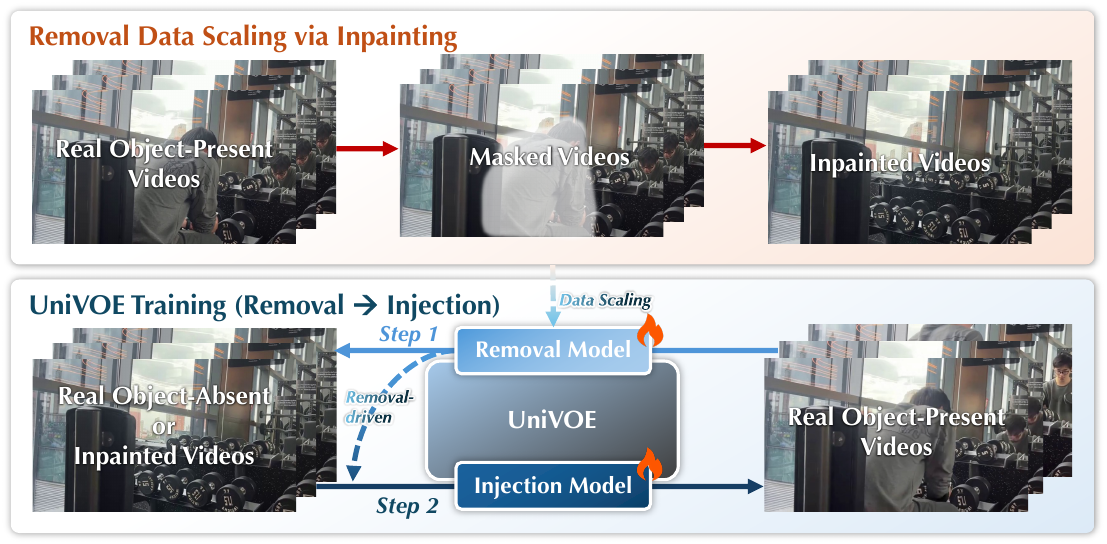}
  \vspace{-1.5em}
  \caption{UniVOE uses removal to scale paired video data up, drives the training of injection.}
  \label{fig:illustration}
  \vspace{-1.em}
\end{figure}

We propose \textbf{UniVOE}, a removal-driven framework for realistic video object removal and injection. As shown in Fig. \ref{fig:illustration}, UniVOE first trains an effect-aware remover by transferring scalable inpainting priors to object removal and refines with limited object-present/absent pairs~\cite{rose,effecterase}. The trained remover is then applied to real object-present videos to derive object-absent counterparts, while the original videos serve as injection targets containing real object motion and interaction effects. These removal-derived pairs, together with limited open-source data, supervise an injection model with flow-matching~\cite{lipman2023flow}. However, pixel-wise supervision alone cannot fully close the perceptual gap of injection: for an object-absent video, multiple appearances, trajectories, and contact patterns may be plausible, while failures such as over-saturation, unnatural motion, missing contact shadows, or unrealistic reflections are better captured by perceptual judgment than by reconstruction error. We therefore introduce an online post-training stage tailored to object injection. It employs a proprietary reward model, finetuned from Qwen3-VL~\cite{qwen3vl}, to rank the motion quality and naturalness of edited videos, steering optimization toward model's performance ceiling. More importantly, as an editing task, the complete frame sequence should be fully covered during ranking and training. To make online post-training feasible for large diffusion models at realistic video resolutions and durations, we design a VRAM-efficient Group-Relative Advantage Regression (GRAR). It avoids the memory overhead of maintaining a reference policy while preserving reward-guided training. Experiments demonstrate that UniVOE achieves SOTA empirical results on both video object removal and injection. As shown in Fig.~\ref{fig:teaser}, for removal, it robustly eliminates target objects together with their shadows, reflections, and contact cues. For injection, it synthesizes objects with plausible motion and interaction effects. Compared with strong open-source and commercial systems, UniVOE is competitive and clearly stronger in challenging interaction-effect cases that prior methods often mishandle. Our contributions are summarized as follows:
\begin{itemize}
    \item We propose an effective training framework to optimize video object removal and injection models with limited real-world paired data, named UniVOE. We exploit their reciprocal-but-asymmetric training structure: removal is first scaled with easier inpainting task and then reused as a supervision engine for the harder injection direction.

    \item To enhance the realism of injection, we develop a novel post-training algorithm, called GRAR, to overcome the limitation of pixel-level reconstruction and align the injected results with human preference. Experiments verify its effectiveness in improving injection quality.

    \item We apply various techniques to removal model for large-scale training data curation. First, object deletion is bootstrapped with video inpainting. Next, we adapt the model with paired data to enable interactive effects removal. Besides, we accelerate inference to facilitate large-scale generation. Finally, we add an effective video filtering mechanism to retain only high-quality data for training.

    \item Extensive comparisons with open-source and commercial systems show that UniVOE achieves SOTA performance on both object removal and injection, especially in challenging interaction-effect erasing and prediction.
\end{itemize}

\section{Related Work}
\label{sec:rw}

\subsection{Video Object Removal}

Video object removal is a special case of video inpainting: the masked region must be filled with plausible and temporally coherent background. Early methods rely on temporal adversarial losses, optical-flow propagation, or spatio-temporal attention to improve completion quality~\cite{chang2019free,zeng2020sttn,liu2021fuseformer,li2022e2fgvi,propainter,controlnet,das,gao2025longvie}, recent diffusion-based pipelines extend the generative range to large holes~\cite{diffueraser,videopainter}. These generic inpainters fill the masked region but are not designed to clean up object-induced effects outside the mask. A second line of work, namely effect-aware removal, targets the shadows, reflections, mirror images, and deformations that an object leaves on the surrounding scene~\cite{rose,objectwiper}. EffectErase~\cite{effecterase} treats object insertion as an auxiliary inverse task to regularize removal. In contrast, we treat removal as a supervision engine for injection: a trained effect-aware remover converts object-present videos into object-absent counterparts for paired supervision of object injection.

\subsection{Video Object Injection}

Object injection places a target object into a host scene while keeping it visually compatible with background. At image level, prior work has explored exemplar-guided editing, object compositing, and zero-shot object-level customization~\cite{yang2023paint,song2023objectstitch,lu2023tficon,chen2024anydoor,Li2025ICCustomDI}. Lifting these to video is harder: the inserted object must remain temporally consistent and move coherently. Recent video editing methods add controllability through first-frame editing, reference-object preservation, or compositing~\cite{ku2024anyv2v,revideo,wei2024dreamvideo,videoanydoor,gencompositor,Unic}, but they typically expect the user to specify object identity, motion track, and other injection-related controls up front. Unlike these methods, UniVOE offers lighter interface (text by default, with an optional reference image). This enables flexible use and large-scale paired data curation.

\subsection{Post-training for Preference Alignment}

Aligning generation results with human perception generally takes one of two routes. Offline preference optimisation~\cite{dpo,xu2023imagereward,prabhudesai2023alignprop,wallace2024diffusiondpo} fine-tunes the model on a fixed set of pairwise preferences. It is simple and stable, but bound to expensive human annotation, and the model never trains on its own samples. Online post-training instead scores samples drawn from the current policy on the fly, which saves manual annotation and allows the model to explore its higher performance boundaries. Within online for diffusion and flow-matching, recent work explores PPO-style trust regions~\cite{liu2025flowgrpo} and negative-aware fine-tuning, exemplified by DiffusionNFT~\cite{zheng2025diffnft}. Although these are effective on image models, applying them off-the-shelf to large video backbones for editing tasks runs into a memory wall: keeping a frozen reference policy alongside a trainable one, plus storing rollouts of long video latents, needs larger VRAM. Therefore, we propose a VRAM-efficient post-training strategy tailored to video editing. It follows a single-pass reward-weighted regression but discards the reference policy, named GRAR.

\begin{figure*}[t]
  \centering
    \includegraphics[width=0.94\linewidth]{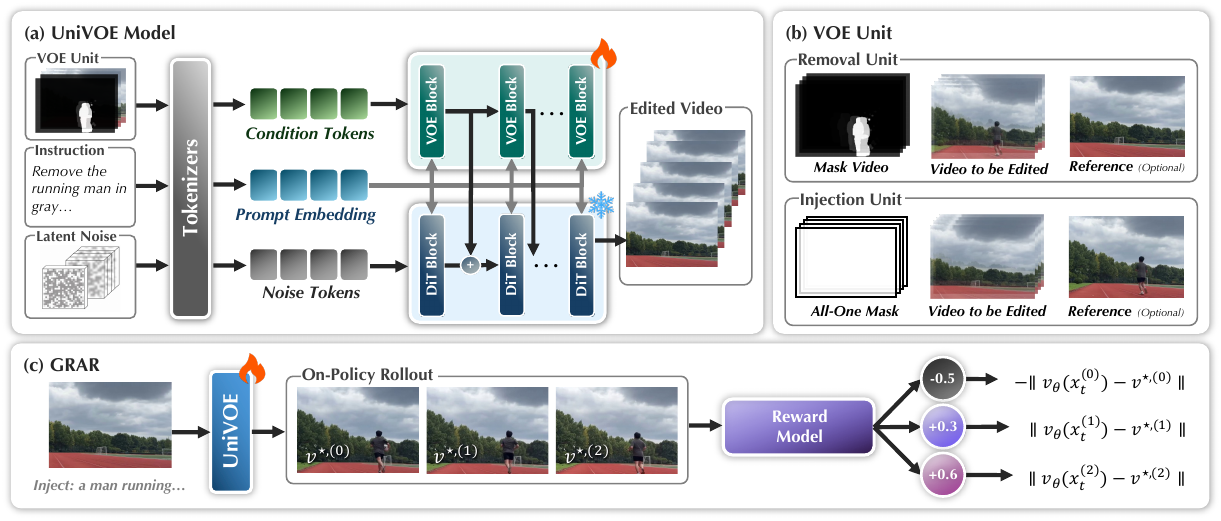}
  \vspace{-2.mm}
  \caption{Workflow of UniVOE. (a) The remover and injector share the same network, but use different instructions and VOE units. (b) For removal, VOE unit contains an input video and an inflated binary mask video to indicate object to be removed. For injection, an all-one mask is used to enable global interaction-aware injection. The reference image is optional in removal and injection. It is generated with external image editing model, providing more specific editing guidance. (c) To improve physical realism of injected results, we propose GRAR to update the pre-trained injection model toward human preferences.}
  \label{fig:workflow}
\end{figure*}

\section{Method}
\label{sec:method}
UniVOE contains two models, an effect-aware remover $\pi_{\rm rem}$ and an injector $\pi_{\rm inj}$, which share the same structure but are trained for opposite directions of object present/absent transformation. Both are built on Wan2.1-14B~\cite{wan21}.

\subsection{Preliminaries: Latent Flow Matching}
\label{sec:method_bg}

Let $x_0\in\mathbb{R}^{C\times T\times H\times W}$ denote the latent of a clean video, and let $\epsilon\sim\mathcal{N}(0,I)$ be Gaussian noise of the same shape. For timestep $t\sim\mathcal{U}[0,1]$, we use the rectified-flow interpolant:
\begin{equation}
\label{eq:interp}
x_t = (1-t)x_0 + t\epsilon,
\qquad
v^\star(x_0,\epsilon,t)=\partial_t x_t=\epsilon-x_0 .
\end{equation}
A network $v_\theta(x_t,t\mid c)$ predicts velocity conditioned on $c$. In our case, $c$ includes an input video, video mask, prompt, and an optional reference image. Flow-matching objective is:
\begin{equation}
\label{eq:fm}
\mathcal{L}_{\rm FM}(\theta)
=
\mathbb{E}_{x_0,\epsilon,t,c}
\left[
\left\|
v_\theta(x_t,t\mid c)-v^\star(x_0,\epsilon,t)
\right\|_2^2
\right].
\end{equation}
At inference, samples are generated by using the learned velocity field from noise to data. Both $\pi_{\rm rem}$ and $\pi_{\rm inj}$ of UniVOE optimize with this objective, but differ in condition $c$.

\subsection{Stage 1: Effect-Aware Video Object Removal}
\label{sec:method_remover}

As Fig.~\ref{fig:illustration} shows, video inpainting and object removal share a common goal: recovering background occluded by target region. This similarity allows us to bootstrap object removal from inpainting data. Further, removal requires erasing object-induced interaction effects, \textit{e.g.}, shadows and reflections that extend beyond object. We train remover in two steps: inpainting pretraining and effect-aware fine-tuning.

\paragraph{Video inpainting.}
We employ SAM2~\cite{ravi2024sam2} to segment foreground objects in real-world videos, obtain their masks $M$. We collect over 63K high-quality videos containing salient objects and apply Gaussian dilation to $M$ to ensure sufficient coverage of object boundaries. In training, we randomly sample two videos, $Vid_A$ and $Vid_B$, and use $M$ from $Vid_B$ to mask corresponding regions in $Vid_A$, producing realistic object-shaped missing regions with temporal variations. The masked $Vid_A$ is used as input, while the original $Vid_A$ serves as the ground truth. The removal model is optimized with the reconstruction loss in Eq.~(\ref{eq:fm}), enabling it to learn a strong scene completion prior for plausibly recovering background content occluded by objects.

\paragraph{Effect-aware fine-tuning.}
An inpainting model fills content inside the masked region, while object removal requires eliminating object-induced effects, such as shadows, reflections, or other interactions that may extend beyond the object mask. To make the remover effect-aware, we further fine-tune it on synthesized and captured object-present/object-absent video pairs collected from prior datasets~\cite{rose,effecterase}, yielding 68K filtered training pairs. During training, we use inflated object mask $M$ that fully covers the target object but leaves interaction effects outside. This encourages the model to recover the complete object-absent scene rather than merely inpainting the object region, while improving robustness to inaccurate masks at inference time. Process is formulated as:
$
\pi_{\rm rem}
\left(
V_{\rm abs}\mid V_{\rm obj}, M, c_{\rm rem}
\right)
$,
where $V_{\rm obj}$ and $V_{\rm abs}$ denote the object-present and object-absent videos, respectively. Removal condition $c_{\rm rem}$ supports both text-only and reference-guided modes:
\begin{equation}
\label{eq:remover_condition}
c_{\rm rem}=
\begin{cases}
p, & \text{without a reference image},\\
(p,I_{\rm ref}), & \text{with a reference image},
\end{cases}
\end{equation}
where $p$ denotes the instruction as shown in Fig.~\ref{fig:workflow}(a), $I_{\rm ref}$ is a reference image specifying the desired object-absent state.

As shown in Fig.~\ref{fig:workflow}(b), the reference-guided mode provides an explicit cue in addition to the input video and mask. Specifically, we edit the first frame of $V_{\rm obj}$ using the same removal instruction $p$ with HunyuanImage3~\cite{hunyuanimage3}, producing $I_{\rm ref}$ in which the target object has been removed. When $I_{\rm ref}$ is provided, UniVOE prepends it to $V_{\rm obj}$ as an additional conditioning frame. Correspondingly, one noise frame is prepended to the latent noise sequence and one all-zero mask frame is prepended to the mask sequence, so that the video, noise, and mask remain temporally aligned. The reference frame provides an explicit but static object-absent prior, guiding the model to remove the dynamic object consistently with the edited first frame throughout the video.

\paragraph{Inference Acceleration.}
When applying remover to large-scale data curation, its inference cost directly limits the scale of paired data we can generate. To accelerate it, we distill remover from the default $50$-steps sampling into $12$-steps. Concretely, we shorten the flow-matching sampler to $12$-steps with a fixed sigma shift of $5.0$, and finetune with Eq.~(\ref{eq:fm}). To synchronize timesteps of the backbone and branch, we add a LoRA~\cite{hu2022lora} of rank $32$ to attention and feed-forward layers of backbone, jointly update the full branch together with LoRA. In inference, we directly remove objects in $12$-steps, making large-scale object-absent video generation practical. Note that our removal results in this paper are all generated from distilled $12$-steps model, which yields a substantial speed-up with imperceptible quality degradation.

\subsection{Stage 2: Removal-Driven Injection Pre-training}
\label{sec:method_injection_sft}
Given a collection of real object-present videos $\{V_{\rm obj}^{(i)}\}$, we apply the trained remover $\pi_{\rm rem}$ to derive paired supervision:
\begin{equation}
\label{eq:derived_pair}
V_{\rm abs}^{(i)}=\pi_{\rm rem}(V_{\rm obj}^{(i)},\tilde{M}^{(i)}).
\end{equation}
The derived video $V_{\rm abs}^{(i)}$ serves as the injection input, while the original video $V_{\rm obj}^{(i)}$ naturally becomes a photorealistic target containing real object motion and interaction effects. Each removal-derived training sample is represented as $\left(V_{\rm abs}^{(i)}, V_{\rm obj}^{(i)}, p^{(i)}, I_{\rm ref}^{(i)}\right)$, where $p^{(i)}$ is the injection prompt and $I_{\rm ref}^{(i)}$ is an optional reference image. We generate object prompts from the target video $V_{\rm obj}^{(i)}$ using Qwen3-VL. As shown in Fig.~\ref{fig:workflow}(b), $I_{\rm ref}^{(i)}$ provides specific appearance and interaction constraints. During training, it can be taken from the first frame of $V_{\rm obj}^{(i)}$. In inference, it is obtained by applying image editing tool~\cite{hunyuanimage3} to the first frame of $V_{\rm abs}^{(i)}$, enabling more explicit injection control.

Totally, we employ removal model to produce 69K usable paired data, combined with the previously filtered 68K open-source data, all 137K video pairs are used to train injection model. We define injection condition as $c_{\rm inj}=(V_{\rm abs},p,I_{\rm ref})$. The injector can be trained with:
\begingroup\small
\begin{equation}
\label{eq:sft}
\mathcal{L}(\theta)
=
\mathbb{E}_{(V_{\rm obj},c_{\rm inj})\sim\mathcal{D}_{\rm inj},\,\epsilon,t}
\left[
\left\|
\left(
v_\theta(x_t,t\mid c_{\rm inj})-v^\star
\right)
\right\|_2^2
\right],
\end{equation}
\endgroup
where $\mathcal{D}_{\rm inj}$ denotes the union of training data. The resulting injection model is denoted as $\pi_{\rm inj}$, which is trained to inject not only the object itself, but also its related realistic interaction effects such as shadows, reflections, contact, \textit{etc}.

\subsection{Stage 3: Post-Training with GRAR}
\label{sec:method_rl}

Although Stage~2 provides basic object injection ability, pixel-wise supervision cannot fully capture perceptual quality. For an object-absent video, multiple appearances, dynamics, and interaction patterns can be plausible, while artifacts such as over-saturation, floating objects, implausible reflections, and incoherent motion are perceptually critical. Hence, we introduce online post-training to improve realism and interaction plausibility.

Given the injection model $\pi_{\rm inj}$, we generate
$V_{\rm gen}\sim\pi_{\theta}^{\rm inj}(\cdot\mid V_{\rm abs},c_{\rm inj})$
and compute a reward $r=Q_{\theta}(V_{\rm gen})$, where $Q_{\theta}$ evaluates object motion and appearance quality. Since native Qwen3-VL cannot reliably assess such qualities, we collect professionally annotated rankings on motion quality and appearance realism and use them to finetune Qwen3-VL into a reward model aligned with human preferences.

In each iteration, we rollout $K$ samples for each condition:
\begin{equation}
\label{eq:rollouts}
\{V_{\rm gen}^{(j)}\}_{j=1}^{K}
\sim
\pi_{\theta_{\rm old}}^{\rm inj}(\cdot\mid V_{\rm abs},c_{\rm inj}),
\end{equation}
and score them through $Q_{\theta}$. Based on the relative advantages within the group, assign positive or negative signs to $K$ scores. We first calculate the group median $\tilde{r}_c$, then assign:
\begin{equation}
\label{eq:adv}
A^{(j)}
=
\begin{cases}
+1, & Q_{\theta}(V_{\rm gen}^{(j)})\ge \tilde{r}_c,\\
-1, & Q_{\theta}(V_{\rm gen}^{(j)})< \tilde{r}_c.
\end{cases}
\end{equation}
This positive or negative sign depends on the in-group ranking, making the update bounded and robust to reward scores.

\paragraph{Advantage regression.}
Post-training methods mainly require double models, a reference policy, a trainable policy, with multiple forward passes. This is because, limited by reward hacking, updates driven solely by rewards cause the model to overfit or collapse into a spurious optimum, far removed from the high-quality distribution learned in pre-training. Hence, reference policy is necessary to constrain post-training scope, but inevitably increases VRAM overhead, limiting its potential for use in larger-scale models. Existing methods are mainly applied to image models or small video backbones such as 1.3B~\cite{pan2025id}. But as to stronger video model such as our injector with 23B parameters, using off-the-shelf solutions is infeasible. One workaround is to reduce VRAM usage by sampling partial frames for training. However, for video editing tasks, abnormal editing effects in any frame are intolerable, \textit{i.e.}, training with full frame sequence is essential for editing. In our case, using 97 frames of 480p leads to unacceptably VRAM load.

To this end, we propose Group-Relative Advantage Regression (GRAR), a single-policy surrogate that preserves preference-aligned optimization without a separate reference model. We rethink the role of reference policy from the nature of video editing. Reference policy constrains the magnitude of post-training for generation. However, for editing, input video itself constrains the output content, \textit{i.e.}, the model does not generate a new video from scratch but merely modifies parts, with the core content directly dictated by input video. Therefore, GRAR only regress single trainable policy with reward values without redundant reference policy.

For each rollout $j$, we treat the generated video as the online sample $x_0^{(j)}$ and recompute its flow-matching target following Eq.~(\ref{eq:interp}). GRAR objective is then formulated as:
\begingroup\small
\begin{equation}
\label{eq:srwfm}
\mathcal{L}_{\rm GRAR}(\theta)
=
\mathbb{E}_{c_{inj},j,t}
\left[
A^{(j)}
\left\|
v_\theta(x_t^{(j)},t\mid c_{inj})-v^{\star,(j)}
\right\|_2^2
\right].
\end{equation}
\endgroup
When $A^{(j)}=+1$, minimizing Eq.~(\ref{eq:srwfm}) pulls the model toward the velocity field of high-reward rollouts. When $A^{(j)}=-1$, the sign reverses the update and pushes the model away from low-reward rollouts. In this way, GRAR provides a negative-aware online update while requiring only a single trainable policy during optimization, making online post-training feasible for large-scale video editing models.

\paragraph{Online training loop.}
At each iteration, we sample a batch of 8 contexts and generate $K=16$ rollouts per context using $\pi_{\theta_{\rm old}}^{\rm inj}$. We score them with $Q_{\theta}$ and compute signed advantages using Eq.~(\ref{eq:adv}). In optimization, only $\pi_{\theta}^{\rm inj}$ is forwarded to minimize Eq.~(\ref{eq:srwfm}), avoiding reference policy storage and forward passes. After each update, we synchronize the rollout policy as $\theta_{\rm old}\leftarrow\theta$. Final policy serves as UniVOE injector.

\begin{figure*}[t]
  \centering
    \includegraphics[width=1.\linewidth]{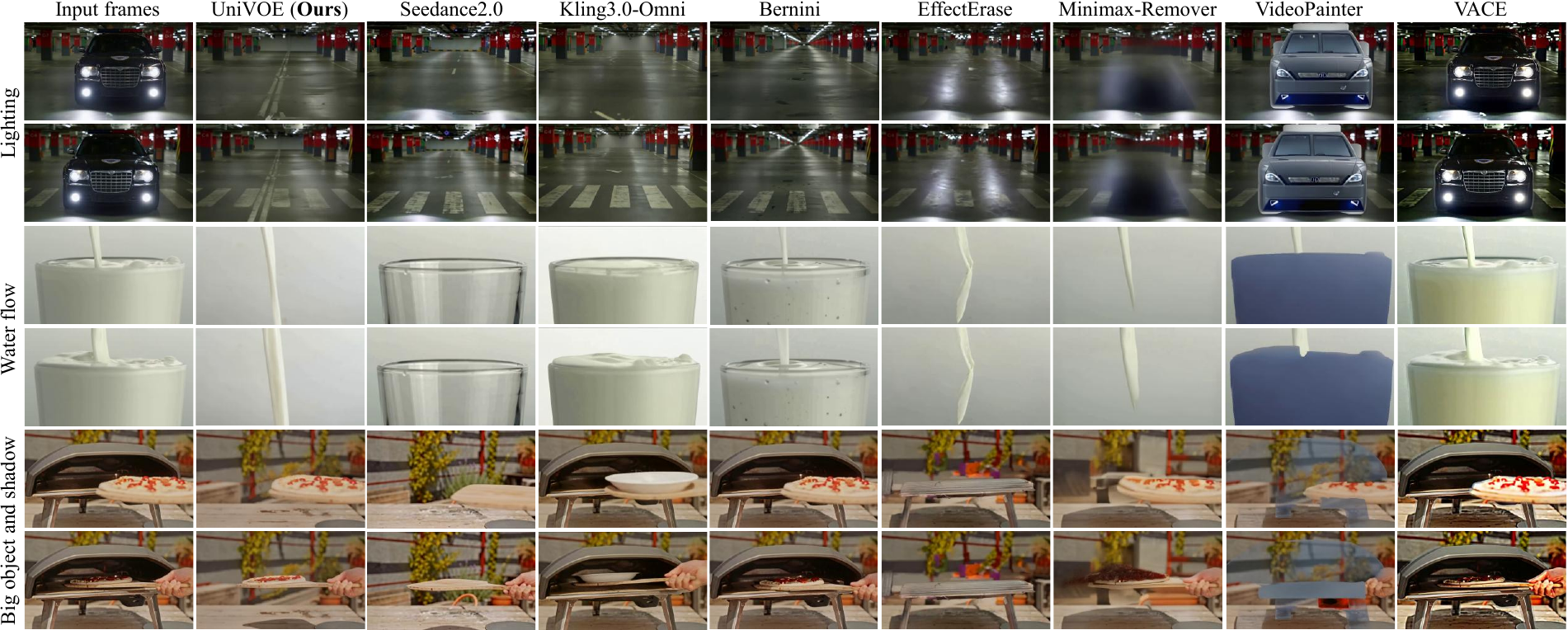}
  \vspace{-1.5em}
  \caption{Visual comparison on video object removal. Other methods either remain object-related effects traces (\textit{e.g.}, lighting of the car), or predict unnatural background contents. In contrast, our method removes object itself and recovers physically realistic scene. It outperforms both open-source and commercial models. More visual comparison can be found in supplementary.}
  \label{fig:comp_rem}
\end{figure*}

\section{Experiments}
\label{sec:exp}
\subsection{Implementation Details}
\label{sec:impde}
Both remover and injector share the same model structure, each contains 23B parameters, including a 14B frozen Transformer backbone, a 3B trainable control branch, and other modules such as VAE, \textit{etc}. To support practical video editing lasting over nine seconds (\textit{i.e.}, more than 200 frames), we train with 97 frames of video clip with resolution of $832\times 480$. Learning rate for remover and injector pre-training is 1e-5, and injector post-training is 1e-6. During inference, both remover and injector support 2 conditioning paradigms: text-only or text-and-reference image. As some compared methods require reference image to inject objects. For fair comparison, here we compare removal with text-only input and compare injection with text-and-reference input.

\begin{table}[t]
\centering
  \resizebox{1.03\linewidth}{!}{
  \begin{tabular}{ll|ccccc|cc|c}
    \toprule
    Data & Metric
    & VidPaint & VACE & Minimax & EffErase & Bernini & Kling3 & Seedance2 & Ours \\
    \hline
    \multirow{4}{*}{VOR}
    & PSNR $\uparrow$  & 17.425 & 11.567 & 17.243 & 23.210 & 16.380 & 21.581 & 20.002 & \textbf{23.752} \\
    & SSIM $\uparrow$  & 0.705  & 0.601  & 0.692  & 0.734  & 0.674 & 0.712 & 0.715 & \textbf{0.786} \\
    & LPIPS $\downarrow$ & 0.253 & 0.441 & 0.229 & 0.142 & 0.247 & 0.182 & 0.204 & \textbf{0.130} \\
    & FVD $\downarrow$ & 563.08 & 825.69 & 171.66 & 150.82 & 238.47 & 210.58 & 174.36 & \textbf{146.37} \\
    \hline
    \multirow{4}{*}{ROSE}
    & PSNR $\uparrow$  & 25.693 & 15.377 & 28.839 & 24.666 & 23.814 & 22.433 & 27.286 & \textbf{29.596} \\
    & SSIM $\uparrow$  & 0.896 & 0.388 & 0.745 & 0.918 & 0.876 & 0.903 & 0.912 & \textbf{0.919} \\
    & LPIPS $\downarrow$ & 0.1205 & 0.3528 & 0.1178 & 0.100 & 0.136 & 0.115 & 0.106 & \textbf{0.098} \\
    & FVD $\downarrow$ & 291.81 & 755.80 & 146.69 & 111.55 & 184.73 & 201.16 & 128.42 & \textbf{85.25} \\
    \bottomrule
  \end{tabular}
  }
  \vspace{-0.5em}
    \caption{Comparison on removal. The best results are highlighted in \textbf{bold}. Video materials are given in supplementary.}
    \label{tab:comp_rem}
\end{table}

\begin{table}[t]
\centering
  \resizebox{1.03\linewidth}{!}{
  \begin{tabular}{ll|cccc|cc|c}
    \toprule
    Data & Metric
    & VidPaint & VACE & UniVideo & Bernini & Kling3 & Seedance2 & Ours \\
    \hline
    \multirow{4}{*}{VOR}
    & DINO-I $\uparrow$  & 0.310 & 0.232 & 0.387 & 0.327 & 0.303 & 0.381 & \textbf{0.390} \\
    & CLIP-P $\uparrow$  & 0.224 & 0.200 & 0.228 & 0.208 & 0.227 & 0.217 & \textbf{0.232} \\
    & LPIPS $\downarrow$ & 0.257 & 0.366 & 0.314 & 0.329 & 0.295 & 0.249 & \textbf{0.237} \\
    & FVD $\downarrow$   & 391.63 & 505.02 & 300.69 & 337.48 & 296.67 & 259.31 & \textbf{234.56} \\
    \hline
    \multirow{4}{*}{ROSE}
    & DINO-I $\uparrow$  & 0.241 & 0.192 & 0.346 & 0.319 & 0.312 & 0.334 & \textbf{0.347} \\
    & CLIP-P $\uparrow$  & 0.220 & 0.217 & 0.236 & 0.242 & 0.221 & 0.243 & \textbf{0.248} \\
    & LPIPS $\downarrow$ & 0.130 & 0.224 & 0.279 & 0.245 & 0.153 & 0.161 & \textbf{0.114} \\
    & FVD $\downarrow$   & 351.42 & 352.24 & 282.01 & 238.72 & 177.63 & 143.56 & \textbf{85.12} \\
    \bottomrule
  \end{tabular}
  }
  \vspace{-0.5em}
    \caption{Comparison on injection. The best results are highlighted in \textbf{bold}. Video materials are given in supplementary.}
    \label{tab:comp_inj}
\end{table}

\begin{figure*}[t]
  \centering
    \includegraphics[width=1\linewidth]{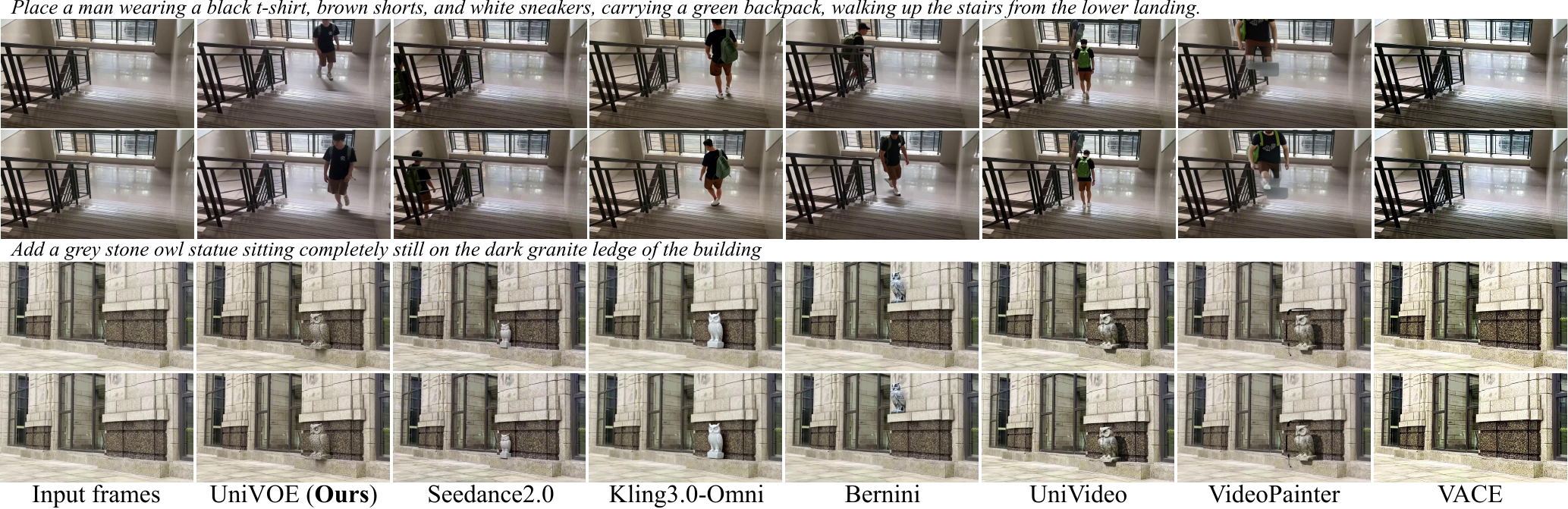}
  \vspace{-1.5em}
  \caption{Visual comparison on video object injection. Corresponding prompt is given above. Other methods either inject background-inconsistent object, or fail to inject harmonious object. Our model predicts matching target and credible dynamics.}
  \label{fig:comp_inj}
\end{figure*}

\subsection{Comparison on Object Removal}
\label{sec:comprem}
We compare removal model with 5 SOTA open-source methods, including VideoPainter (SIGGRAPH 2025)~\cite{videopainter}, VACE (ICCV 2025)~\cite{vace}, Minimax-Remover (NeurIPS 2026)~\cite{minimax}, EffectErase (CVPR 2026)~\cite{effecterase}, and Bernini~\cite{bernini}, and commercial models Kling3-Omni~\cite{kling} and Seedance2.0~\cite{seedance2}. Note that we only input text prompt to the removal model. As shown in Fig.~\ref{fig:comp_rem}, we compare removal performance on various interaction effects, \textit{e.g.}, heavy occlusion and obvious lighting. One can see that among all State-Of-The-Art (SOTA) methods, our method realizes the best object removal and effects recovery. Especially for the bottom cases, where the glass and oven are removed, respectively. For glass case, our model recovers naturally flowing milk but others fail. For oven case, other methods all remain the shadow of removed oven, resulting sub-optimal recovered background effected by it. In contrast, UniVOE precisely recovers a pizza in direct sunlight and its shadow. We believe this is because UniVOE is firstly trained from massive inpainting data, which brings strong object deletion and background recovery priors.

For quantitative evaluation, we compare on two benchmarks: ROSE containing 56 pairs, and VOR containing 100 pairs. Note that test sets are excluded from the training data. We employ fidelity metrics to compare objectively, including PSNR~\cite{psnr}, SSIM~\cite{ssim}, LPIPS~\cite{lpips}, and FVD~\cite{fvd}. As shown in Tab.~\ref{tab:comp_rem}, our method achieves the best objective metrics across benchmarks, which means it removes objects and effects, and recovers the most authentic background. This is consistent with subjective perception, further proves the effectiveness of the proposed new training method, \textit{i.e.}, starting removal training from video inpainting.

\begin{figure}[t]
  \centering
    \includegraphics[width=1\linewidth]{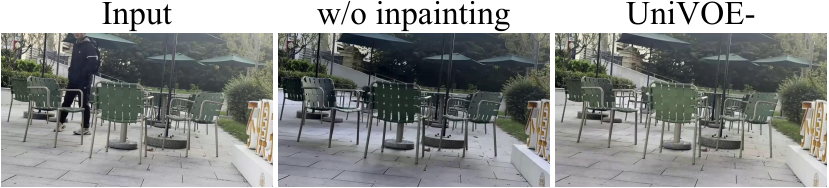}
  \vspace{-1.5em}
  \caption{Visual ablation of training data used for the remover. ``w/o inpainting'' is trained by noisy removal pairs and carries the risk of overall color shifts. ``UniVOE-'' introduces inpainting data to avoid excessive background editing.}
  \label{fig:abla_rem}
\end{figure}

\subsection{Comparison on Object Injection}
\label{sec:compinj}
We compare UniVOE-injector with 4 open-source methods: VideoPainter, VACE, UniVideo (ICLR 2026)~\cite{univideo}, and Bernini~\cite{bernini}, and commercial models, Kling3-Omni and Seedance2.0. Similar to UniVideo, our injector is also a mask-free approach, capable of injecting based on prompts without explicit mask inputs. This setting is more challenging and practical for large-scale batch video injection. As shown in Fig.~\ref{fig:comp_inj}, our method generates the most realistic injected results. VideoPainter, VACE, and UniVideo fail to place realistic objects. Although Kling3 and Seedance2 can produce competitive results, our outputs show more harmonious appearances and natural dynamics. Source videos can be found in supplementary for intuitive illustration.

We conduct quantitative comparison on the same test sets used before. Similar to previous injection method~\cite{univideo}, we employ LPIPS, FVD, DINO-I~\cite{dino}, and CLIP-P, where DINO-I aims to measure identity consistency, CLIP-P measures prompt following. As shown in Tab.~\ref{tab:comp_inj}, our method outperforms all algorithms compared across benchmarks, which proves our superior injection ability and is consistent with visual results.

\begin{table}[t]
\centering
  \resizebox{\linewidth}{!}{
  \begin{tabular}{l|cc|cccc}
    \toprule
    Settings & Inpaint & Removal & PSNR $\uparrow$ & SSIM $\uparrow$ & LPIPS $\downarrow$ & FVD $\downarrow$ \\
    \hline
    w/o inpainting & 0K  & 68K & 20.578 & 0.711 & 0.158 & 163.130 \\
    UniVOE-       & 34K & 34K & 22.891 & 0.740 & 0.141 & 149.777 \\
    \bottomrule
  \end{tabular}
  }
  \vspace{-0.5em}
  \caption{Ablation on inpainting data for training remover.}
  \label{tab:ablation_rem}
\end{table}

\begin{table}[t]
\centering
  \resizebox{\linewidth}{!}{
  \begin{tabular}{l|cccc}
    \toprule
    Settings & DINO-I $\uparrow$ & CLIP-P $\uparrow$ & LPIPS $\downarrow$ & FVD $\downarrow$ \\
    \hline
    w/o GRAR & 0.353 & 0.221 & 0.307 & 361.286 \\
    UniVOE & 0.390 & 0.232 & 0.237 & 234.562 \\
    \bottomrule
  \end{tabular}
  }
  \vspace{-0.5em}
  \caption{Ablation on post-training stage for object injection.}
  \label{tab:ablation_inj}
\end{table}

\subsection{Ablation Study}
\label{sec:ablas}
\paragraph{Scaling up removal data through inpainting.}
We realize SOTA removal ability by bootstrapping with inpainting. We compare two ablation settings to support this claim: (1) ``UniVOE-'', a degraded model trained on 68K hybrid data, including 34K inpainting pairs and 34K removal pairs. (2) ``w/o inpainting'' that is totally trained on 68K removal pairs. Note that UniVOE- only uses part of inpainting and removal pairs to train, which aims to maintain the same data scale as ``w/o inpainting'' for fair comparison. Quantitative comparison is given in Tab.~\ref{tab:ablation_rem}. Although ``UniVOE-'' is degraded by retraining on smaller data scale, it still outperforms ``w/o inpainting'' across all metrics. Visual results in Fig.~\ref{fig:abla_rem} show that ``w/o inpainting'' may shift overall tone. This is because the paired object-present/absent pairs capture in the wild cannot ensure pixel-level background consistency, using such noisy data to train misleads the model to over-edit, changing overall pixels even they are actually not related to the removed object. While inpainting data guaranties strict pixel-level correspondence, improving background preservation ability.

\paragraph{Ablation on post-training stage.}
To validate the positive impact of post-training for video object injection, we compare performance between ``UniVOE'' and the ablation setting ``w/o GRAR'', in which the injector that is only pre-trained. As shown in Tab.~\ref{tab:ablation_inj}, ``UniVOE'' stably outperforms ``w/o GRAR'' across all metrics. Due to limited space of main paper, related visual comparison is given in supplementary.

\section{Conclusion}
\label{sec:conclusion}

We presented UniVOE, a removal-driven framework for realistic video object removal and injection. Our key observation is that the two tasks share the same object-present/object-absent training structure, but differ in learning difficulty: removal can be easily bootstrapped from scalable video inpainting, whereas injection requires strict data pairs for object motion, contact, and interaction effects. UniVOE turns this reciprocal-but-asymmetric structure into a new training strategy. It first learns a strong effect-aware remover, then uses the remover to derive object-absent counterparts from real object-present videos, making the original videos serve as supervision for injection. To further address the perceptual ambiguity left by reconstruction loss, UniVOE introduces an online reward-based post-training stage and Group-Relative Advantage-Regression (GRAR), enabling VRAM-efficient preference-aligned optimization for large video diffusion backbones. Extensive experiments show that UniVOE achieves SOTA performance on object editing, with clear advantages on challenging interaction-effect cases such as shadows, reflections, and motion-dependent visual changes.

\bibliography{aaai2027}

\clearpage
\appendix
\setcounter{secnumdepth}{2}
\twocolumn[{\centering\Large\bfseries Supplementary Material\par\vspace{1em}}]
\section{Video Filtering Gate for Removal Counterparts}
\label{supp:sec:filter}

As described in the main paper, our injector is pre-trained on removal-derived pairs: we run the effect-aware remover $\pi_{\rm rem}$ on real object-present videos $V_{\rm obj}$ to obtain object-absent counterparts $V_{\rm abs}$, and reverse the direction so that $V_{\rm obj}$ becomes a photorealistic injection target. This bootstrap is only as clean as the remover output. A pair in which the remover left a residual object, a hole, or a ghosting artifact, or in which it corrupted the background outside the object region, is worse than useless as injection supervision: reversing such a pair teaches the injector to hallucinate incomplete objects or to repaint background that should stay fixed. We therefore pass every removal-derived pair through an automatic filtering gate before it enters the injection training set $\mathcal{D}_{\rm inj}$, and apply the same gate to open-source object-present/absent pairs~\cite{effecterase}. This section details that gate.

\paragraph{Object region.}
The gate is region-aware: it distinguishes the object region (where the object must disappear) from the background (which must be preserved). For each clip we localize the target with GroundingDINO on the first frame, propagate the box through SAM2 to obtain a per-frame instance mask, and take the union over all target objects into a single binary mask. The mask is softly dilated (Gaussian blur followed by a low threshold) so that the ``inside'' region covers the object together with a margin for its immediate interaction effects, and the complement defines the background region used for the fidelity check below.

\paragraph{Two acceptance criteria.}
A pair is accepted only if it passes both of the following, mirroring the two failure modes of removal:
\begin{itemize}
\item \textbf{Background fidelity (outside the dilated mask).} The remover must not over-edit the scene. We measure the residual change between $V_{\rm abs}$ and $V_{\rm obj}$ restricted to the background region; a pair is rejected when this residual is large, indicating that the remover altered content unrelated to the target (e.g., a global color shift or repainted structure).
\item \textbf{Removal completeness (inside the mask).} The target must be genuinely gone, not merely faded, partially erased, displaced, or replaced by a look-alike. We assess the object region by visual-semantic comparison between $V_{\rm abs}$ and $V_{\rm obj}$; any surviving trace of the target causes rejection.
\end{itemize}

\paragraph{VLM-judge implementation.}
Because ``a surviving trace'' and ``an unnatural background'' are perceptual rather than pixel-exact, both criteria are adjudicated by a vision-language judge (Qwen3-VL) run at temperature $0$ for determinism. The judge is prompted to (i) re-caption both videos and enumerate every object present in $V_{\rm obj}$ but absent in $V_{\rm abs}$, not only the one named in the removal instruction, since the remover sometimes cleanly deletes co-occurring objects, (ii) list the actual changes, (iii) run an explicit residual check for any leftover trace of the target, and (iv) emit a binary decision with a short reason. The instruction that produced the clip is provided only as the intended edit; the judge is told to compare the two videos independently rather than trust the instruction, so that imperfect removals are not waved through. The decision rule is strict-by-default: when the judge is uncertain whether the object is fully gone, it rejects.

\paragraph{Focused residual gate on the object region.}
Full-frame video judgment tends to overlook small residuals that appear only in the middle or late frames. To harden the completeness check, we add a second, focused pass that operates only on the object region: we crop the dilated mask bounding box from $V_{\rm obj}$ and $V_{\rm abs}$, tile eight uniformly sampled frames of each into a static grid, and ask the judge to inspect the result grid cell by cell. If any cell still shows the target --- complete, partial, a faded outline, or a similar substitute --- the pair is rejected. Static grids expose per-frame detail that a played-back video dilutes, making this pass stricter than the video-level judgment; a pair must clear both to be retained.

\paragraph{Deployment and yield.}
The gate is embedded in the multi-node removal pipeline: each node runs the remover and the judge over a shard of clips, with many judge queries issued concurrently, so filtering keeps pace with generation across the cluster. After applying the gate, we retain 69K high-quality removal-derived pairs, which are combined with 68K gate-filtered open-source pairs to form the injection pre-training set $\mathcal{D}_{\rm inj}$.

\section{Pseudo Algorithm for GRAR}
Algorithm~\ref{supp:alg:grar} details the GRAR online post-training loop. Each iteration first rolls out $K$ candidates per context from the frozen rollout policy $\pi_{\theta_{\rm old}}^{\rm inj}$ and scores them with the human-aligned reward model $Q_{\theta}$. A $\{+1,-1\}$ signed advantage is then assigned by comparing each score against the group median $\tilde{r}_{c}$, which keeps the update bounded and insensitive to the absolute reward scale. Finally, each rollout is re-used as an online flow-matching sample and the single trainable policy $\pi_{\theta}^{\rm inj}$ is regressed toward ($A{=}{+}1$) or away from ($A{=}{-}1$) its own velocity target. No reference policy is stored or forwarded, so the only network instantiated during optimization is $\pi_{\theta}^{\rm inj}$ itself, making full-sequence (97-frame, 480p) post-training of the 23B injector feasible within an 80\,GB VRAM budget.

\begin{algorithm}[t]
\caption{Group-Relative Advantage Regression (GRAR) --- Stage~3 online post-training}
\label{supp:alg:grar}
\begin{algorithmic}[1]
\REQUIRE Stage-2 injector $\pi_{\theta}^{\rm inj}$; reward model $Q_{\theta}$; context pool $\mathcal{C}=\{(V_{\rm abs},c_{\rm inj})\}$; group size $K{=}16$; contexts per iter $B{=}8$; VAE encoder $\mathcal{E}$; learning rate $\eta$; iterations $N$
\STATE $\theta_{\rm old}\leftarrow\theta$ \COMMENT{rollout policy initialized from current policy}
\FOR{$n=1$ \textbf{to} $N$}
  \STATE sample $B$ contexts $\{(V_{\rm abs}^{(i)},c_{\rm inj}^{(i)})\}_{i=1}^{B}\sim\mathcal{C}$;\quad $\mathcal{D}\leftarrow\varnothing$
  \STATE \textbf{// (a) Rollout \& reward scoring (no gradient)}
  \FOR{each context $i=1$ \textbf{to} $B$}
     \FOR{$j=1$ \textbf{to} $K$}
        \STATE $V_{\rm gen}^{(i,j)}\sim\pi_{\theta_{\rm old}}^{\rm inj}(\cdot\mid V_{\rm abs}^{(i)},c_{\rm inj}^{(i)})$ \COMMENT{multi-step sampler, frozen $\theta_{\rm old}$}
        \STATE $r^{(i,j)}\leftarrow Q_{\theta}(V_{\rm gen}^{(i,j)})$
     \ENDFOR
     \STATE \textbf{// (b) Group-relative signed advantage \hfill Eq.~(\ref{supp:eq:adv})}
     \STATE $\tilde{r}_{c}^{(i)}\leftarrow \mathrm{median}\{r^{(i,1)},\dots,r^{(i,K)}\}$
     \FOR{$j=1$ \textbf{to} $K$}
        \STATE $A^{(i,j)}\leftarrow +1$ \textbf{if} $r^{(i,j)}\ge \tilde{r}_{c}^{(i)}$ \textbf{else} $-1$
        \STATE $x_0^{(i,j)}\leftarrow \mathcal{E}(V_{\rm gen}^{(i,j)})$ \COMMENT{treat the rollout as the online sample}
        \STATE $\mathcal{D}\leftarrow\mathcal{D}\cup\{(x_0^{(i,j)},\,c_{\rm inj}^{(i)},\,A^{(i,j)})\}$
     \ENDFOR
  \ENDFOR
  \STATE \textbf{// (c) Single-policy advantage regression \hfill Eq.~(\ref{supp:eq:srwfm})}
  \FOR{minibatch $(x_0,c_{\rm inj},A)\subset\mathcal{D}$}
     \STATE $t\sim p(t)$,\quad $\epsilon\sim\mathcal{N}(0,\mathbf{I})$
     \STATE $x_t\leftarrow(1-t)\,x_0+t\,\epsilon$,\quad $v^{\star}\leftarrow \epsilon-x_0$ \COMMENT{flow-matching target, Eq.~(\ref{supp:eq:interp})}
     \STATE $\mathcal{L}_{\rm GRAR}\leftarrow A\,\|v_{\theta}(x_t,t\mid c_{\rm inj})-v^{\star}\|_2^{2}$
     \STATE $\theta\leftarrow\theta-\eta\,\nabla_{\theta}\mathcal{L}_{\rm GRAR}$ \COMMENT{only $\pi_{\theta}^{\rm inj}$ is forwarded; no reference policy}
  \ENDFOR
  \STATE $\theta_{\rm old}\leftarrow\theta$ \COMMENT{synchronize rollout policy}
\ENDFOR
\RETURN $\pi_{\theta}^{\rm inj}$ \COMMENT{final UniVOE injector}
\end{algorithmic}
\end{algorithm}

\section{Compare to DiffusionNFT}
\label{supp:sec:cmp_nft}

GRAR shares the negative-aware spirit of DiffusionNFT~\cite{zheng2025diffnft}, an online post-training method for flow-matching models, but removes its frozen reference policy. In this section we make the relation precise. We first restate the GRAR objective in a self-contained form, then show that (i) the GRAR gradient is first-order consistent with the reward policy gradient (Proposition~2), and (ii) the GRAR update coincides with the negative-anchored DiffusionNFT update up to a first-order residual in the step size (Lemma~1), while instantiating only a single trainable network. We close with the memory argument and the $1.3$B empirical comparison.

\paragraph{Setup and notation.}
Let $\pi_\theta^{\rm inj}(\cdot\mid V_{\rm abs},c_{\rm inj})$ be the injector and $Q_\theta$ the reward model, and write $J(\theta)=\mathbb{E}_{V_{\rm gen}\sim\pi_\theta^{\rm inj}}[Q_\theta(V_{\rm gen})]$ for the expected reward. For a rollout latent $x_0$, noise $\epsilon\sim\mathcal{N}(0,\mathbf{I})$, and $t\sim\mathcal{U}[0,1]$, the rectified-flow interpolant and target are
\begin{equation}
\label{supp:eq:interp}
x_t=(1-t)\,x_0+t\,\epsilon,
\qquad
v^{\star}=\epsilon-x_0 .
\end{equation}
For each context we draw $K$ rollouts $\{x_0^{(j)}\}$ from the frozen policy $\pi_{\theta_{\rm old}}^{\rm inj}$, score them $r^{(j)}=Q_\theta(x_0^{(j)})$, and assign the median-split signed advantage
\begin{equation}
\label{supp:eq:adv}
A^{(j)}=
\begin{cases}
+1, & r^{(j)}\ge \tilde{r}_c,\\
-1, & r^{(j)}< \tilde{r}_c,
\end{cases}
\qquad
\tilde{r}_c=\mathrm{median}\{r^{(j)}\}.
\end{equation}
The GRAR objective is
\begin{equation}
\label{supp:eq:srwfm}
\mathcal{L}_{\rm GRAR}(\theta)
=
\mathbb{E}_{c_{\rm inj},j,t}
\big[
A^{(j)}\,\lVert v_\theta(x_t^{(j)},t\mid c_{\rm inj})-v^{\star,(j)}\rVert_2^2
\big].
\end{equation}
We write $\eta$ for the SGD step size, and $\theta_{\rm old}$ for the rollout parameters, which are synchronized to $\theta$ after every update, so that $\lVert\theta-\theta_{\rm old}\rVert=O(\eta)$ within an iteration.

\paragraph{Bounded, zero-mean advantage.}
\noindent\textbf{Proposition 1 (bounded, zero-mean advantage).} The signed advantage in Eq.~(\ref{supp:eq:adv}) satisfies $|A^{(j)}|=1$; and if the group rewards are i.i.d.\ from an absolutely continuous distribution, then $\mathbb{E}[A^{(j)}]=0$.

\noindent\textit{Proof.} $|A^{(j)}|=1$ is immediate from Eq.~(\ref{supp:eq:adv}). For continuous rewards, ties occur with probability zero, so exactly half of the group lies above the median; the numbers of $+1$ and $-1$ advantages are therefore equal and $\mathbb{E}[\sum_j A^{(j)}]=0$, whence $\mathbb{E}[A^{(j)}]=0$. \hfill$\square$

Boundedness keeps the update insensitive to the numerical scale of $Q_\theta$, and zero mean makes $A^{(j)}$ a valid (baseline-subtracted) advantage weight.

\paragraph{GRAR as a policy-gradient surrogate.}
The flow-matching evidence lower bound~\cite{lipman2023flow} gives $\log\pi_\theta(x_0\mid c)\ge -\mathcal{L}_{\rm FM}(x_0,\theta)+C(x_0,c)$, where $\mathcal{L}_{\rm FM}(x_0,\theta)=\mathbb{E}_{t,\epsilon}\lVert v_\theta-v^{\star}\rVert_2^2$ and $C$ is independent of $\theta$; hence $\nabla_\theta\log\pi_\theta\approx-\nabla_\theta\mathcal{L}_{\rm FM}$.

\noindent\textbf{Proposition 2 (surrogate policy gradient).} Assume $\theta\mapsto v_\theta$ is continuously differentiable and the reward is bounded. Then
\begingroup\scriptsize
\begin{equation}
\label{supp:eq:pg-equiv}
\nabla_\theta\mathcal{L}_{\rm GRAR}(\theta)\big|_{\theta=\theta_{\rm old}}
=
-\nabla_\theta J(\theta)\big|_{\theta=\theta_{\rm old}}
+\mathcal{R},
\quad
\lVert\mathcal{R}\rVert=O(\lVert\theta-\theta_{\rm old}\rVert).
\end{equation}
\endgroup
Consequently, a descent step on $\mathcal{L}_{\rm GRAR}$ is, to first order, an ascent step on the expected reward $J$.

\noindent\textit{Proof.} The advantage policy gradient is $\nabla_\theta J=\mathbb{E}_{x_0\sim\pi_\theta}[A\,\nabla_\theta\log\pi_\theta]$, valid because the latent support (full-dimensional Gaussian noise) does not depend on $\theta$ and any $c$-only baseline may be subtracted from the zero-mean score. Substituting $\nabla_\theta\log\pi_\theta=-\nabla_\theta\mathcal{L}_{\rm FM}+\nabla_\theta g$, with $g\ge0$ the ELBO gap, yields $\nabla_\theta J=-\mathbb{E}_{x_0\sim\pi_\theta}[A\,\nabla_\theta\mathcal{L}_{\rm FM}]+\mathbb{E}[A\,\nabla_\theta g]$. Fixing the sampling distribution to $\pi_{\theta_{\rm old}}$, the first term equals $-\nabla_\theta\mathcal{L}_{\rm GRAR}$; replacing $\pi_\theta$ by $\pi_{\theta_{\rm old}}$ incurs an off-policy bias of order $\lVert\theta-\theta_{\rm old}\rVert$ through the importance ratio, and $\mathbb{E}[A\,\nabla_\theta g]$ is a variational residual that is $\theta$-independent to leading order. Collecting both into $\mathcal{R}$ gives Eq.~(\ref{supp:eq:pg-equiv}). \hfill$\square$

\paragraph{First-order relation to DiffusionNFT.}
DiffusionNFT anchors the update to a frozen reference $v_{\theta_{\rm old}}$: its positive branch regresses $v_\theta$ toward $v^{\star}$ (as GRAR does for $A^{(j)}=+1$), whereas its negative branch regresses toward the anchored target~\cite{zheng2025diffnft}
\begin{equation}
\label{supp:eq:nft-target}
v^{\star}_{\rm neg}
:=
v_{\theta_{\rm old}}+\lambda\,(v_{\theta_{\rm old}}-v^{\star}),
\qquad
\lambda>0,
\end{equation}
which requires an extra forward pass through $v_{\theta_{\rm old}}$ at every gradient step (\citet{zheng2025diffnft} use $\lambda=1$).

\noindent\textbf{Lemma 1 (first-order equivalence).} Let $v_\theta$ be $L_v$-Lipschitz in $\theta$ with $\lVert\theta-\theta_{\rm old}\rVert\le \eta L_v$. Then
\begin{equation}
\label{supp:eq:nft-equiv}
\lVert v_\theta-v^{\star}_{\rm neg}\rVert_2^2
=
-\lambda\,\lVert v_\theta-v^{\star}\rVert_2^2
+B(\theta_{\rm old})
+O(\eta),
\end{equation}
where $B(\theta_{\rm old})$ does not depend on $\theta$. Hence $\nabla_\theta\lVert v_\theta-v^{\star}_{\rm neg}\rVert_2^2=-\lambda\,\nabla_\theta\lVert v_\theta-v^{\star}\rVert_2^2+O(\eta)$.

\noindent\textit{Proof.} Let $\Delta:=v_\theta-v_{\theta_{\rm old}}$, so $\lVert\Delta\rVert\le L_v\lVert\theta-\theta_{\rm old}\rVert=O(\eta)$, and $u:=v_{\theta_{\rm old}}-v^{\star}$, which is independent of $\theta$. From Eq.~(\ref{supp:eq:nft-target}), $v_\theta-v^{\star}_{\rm neg}=\Delta-\lambda u$, so
\begin{equation}
\label{supp:eq:nft-expand}
\lVert v_\theta-v^{\star}_{\rm neg}\rVert_2^2
=\lVert\Delta\rVert^2-2\lambda\langle\Delta,u\rangle+\lambda^2\lVert u\rVert^2 .
\end{equation}
Expanding $\lVert v_\theta-v^{\star}\rVert_2^2=\lVert u+\Delta\rVert^2=\lVert u\rVert^2+2\langle\Delta,u\rangle+\lVert\Delta\rVert^2$ gives $2\langle\Delta,u\rangle=\lVert v_\theta-v^{\star}\rVert^2-\lVert u\rVert^2-\lVert\Delta\rVert^2$. Substituting into Eq.~(\ref{supp:eq:nft-expand}),
\[
\lVert v_\theta-v^{\star}_{\rm neg}\rVert_2^2
=-\lambda\,\lVert v_\theta-v^{\star}\rVert_2^2
+(\lambda+\lambda^2)\lVert u\rVert^2
+(1+\lambda)\lVert\Delta\rVert^2 .
\]
The middle term depends only on $\theta_{\rm old}$ (set it to $B(\theta_{\rm old})$); the last is $O(\eta^2)$. Differentiating in $\theta$ removes $B$ and leaves an $O(\eta)$ gradient residual. \hfill$\square$

Because the positive branch is identical in the two losses and the negative branch agrees up to Eq.~(\ref{supp:eq:nft-equiv}), the DiffusionNFT gradient and the GRAR gradient---with $A^{(j)}=\pm1$ selecting the branch and $\lambda$ absorbed into the learning rate---coincide to first order in $\eta$. Under the per-iteration synchronization $\theta_{\rm old}\!\leftarrow\!\theta$ and the small step size used here, the reference lags the policy by $O(\eta)$, so this first-order agreement holds throughout training and the accumulated trajectory deviation stays within the mini-batch SGD noise floor. GRAR thus realizes the same negative-anchored update while dropping the frozen reference forward pass.

\paragraph{Memory and feasibility.}
The practical consequence of Lemma~1 is that GRAR removes the second network. DiffusionNFT must keep $v_\theta$ (trainable, with optimizer states and activations) co-resident with a frozen $v_{\theta_{\rm old}}$ evaluated by an additional forward pass, roughly doubling the resident model memory. At our $23$B injector scale with full-sequence $97$-frame $480$p inputs, this exceeds the $80$\,GB budget, whereas GRAR instantiates only $\pi_\theta^{\rm inj}$ and remains feasible. This is why prior online post-training is largely restricted to image models or small ($\sim$1.3B) video backbones~\cite{pan2025id}.

\begin{table}[t]
\centering
  \caption{Post-training ablation for video object injection. Evaluate on VOR.}
  \vspace{-1.0em}
  \label{supp:tab:ablation_inj}
  \resizebox{\linewidth}{!}{
  \begin{tabular}{l|cccc}
    \toprule
    Settings & DINO-I $\uparrow$ & CLIP-P $\uparrow$ & LPIPS $\downarrow$ & FVD $\downarrow$ \\
    \hline
    w/ DiffusionNFT & 0.388 & 0.229 & 0.289 & 314.101 \\
    w/ GRAR       & 0.387 & 0.231 & 0.287 & 283.812 \\
    \bottomrule
  \end{tabular}
  }
  \vspace{-1.0em}
\end{table}

\begin{figure*}[t]
  \centering
  \includegraphics[width=0.98\textwidth]{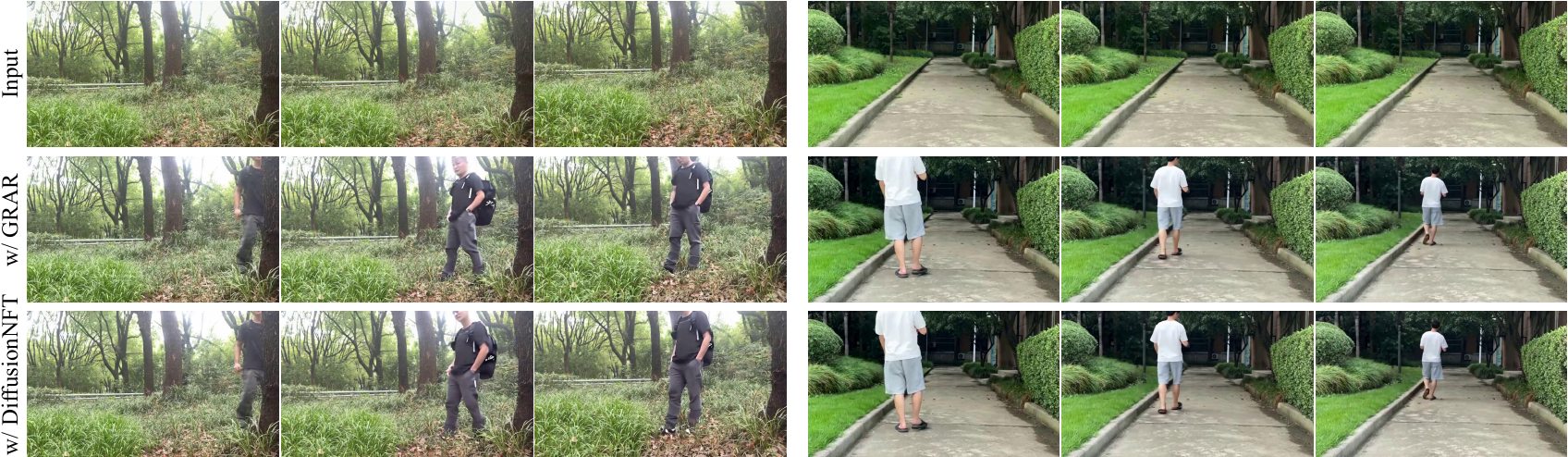}
  \caption{Visual comparison of models trained by GRAR and DiffusionNFT. One can see that their outputs are visually consistent, which is consistent with the quantitative results in Tab.~\ref{supp:tab:ablation_inj}. This further demonstrates that the performance of the proposed GRAR approximates that of DiffusionNFT.}
  \label{supp:fig:supp_rl}
\end{figure*}

\begin{figure}[!t]
  \centering
  \includegraphics[width=0.95\linewidth]{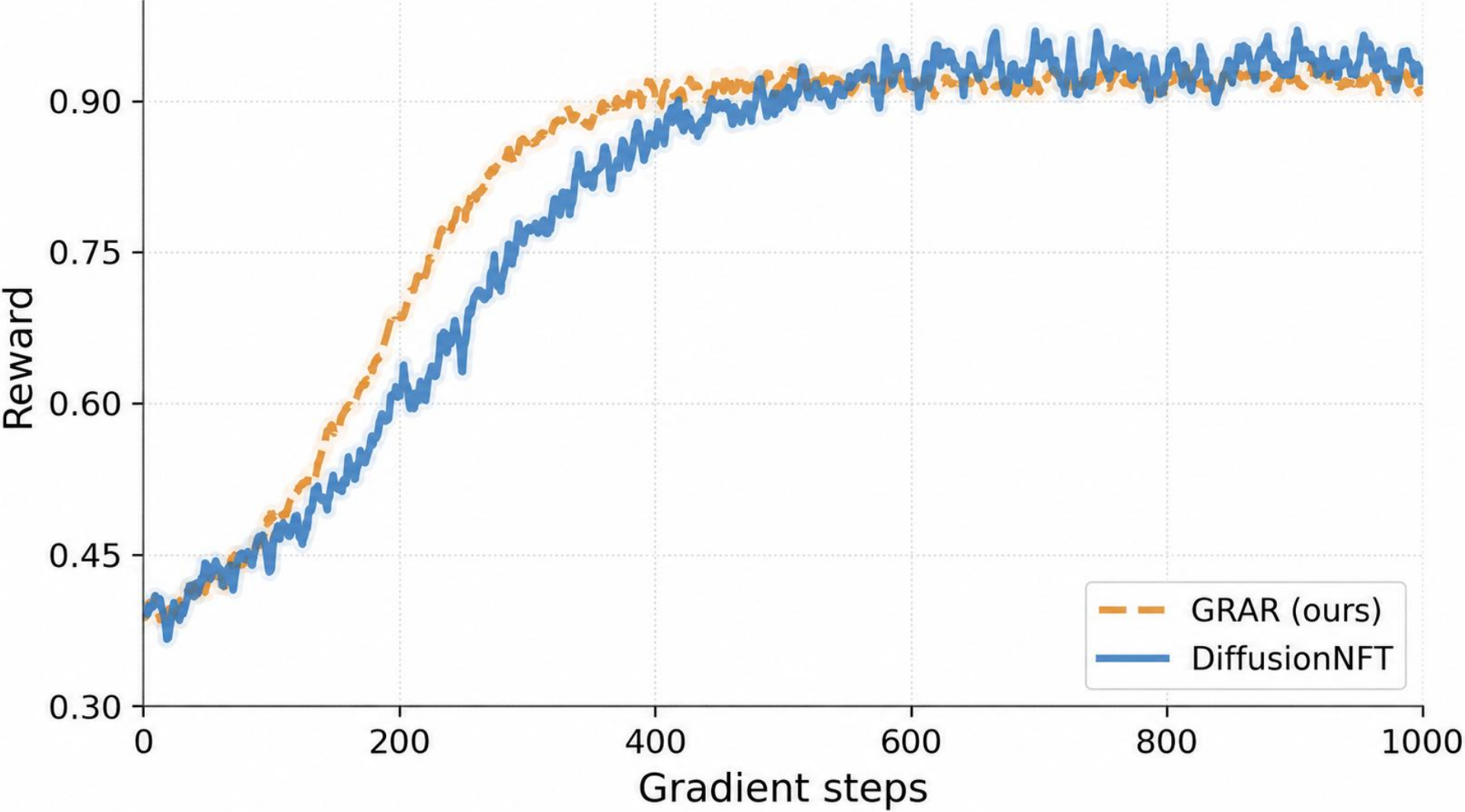}
  \caption{Reward curves of GRAR and DiffusionNFT on the smaller injectors with Wan2.1-1.3B backbone.}
  \label{supp:fig:reward_curve}
\end{figure}

\paragraph{Empirical comparison at 1.3B.}
Since strict DiffusionNFT is infeasible at $23$B, we verify the first-order equivalence at a scale where both fit in memory: a $1.3$B injector pre-trained on the same paired data and then post-trained with GRAR and with strict DiffusionNFT under matched rollouts, reward model, and schedule. The two obtain closely matching quantitative scores (Tab.~\ref{supp:tab:ablation_inj}, lower block) and closely overlapping reward curves (Fig.~\ref{supp:fig:reward_curve}), with qualitatively indistinguishable injections (Fig.~\ref{supp:fig:supp_rl}), confirming that the single-policy GRAR update tracks the DiffusionNFT trajectory in practice. At the full $23$B scale, GRAR is the only feasible option under the $80$\,GB budget, and we therefore adopt it as the default post-training objective for the UniVOE injector.

\begin{figure*}[t]
  \centering
    \includegraphics[width=1\linewidth]{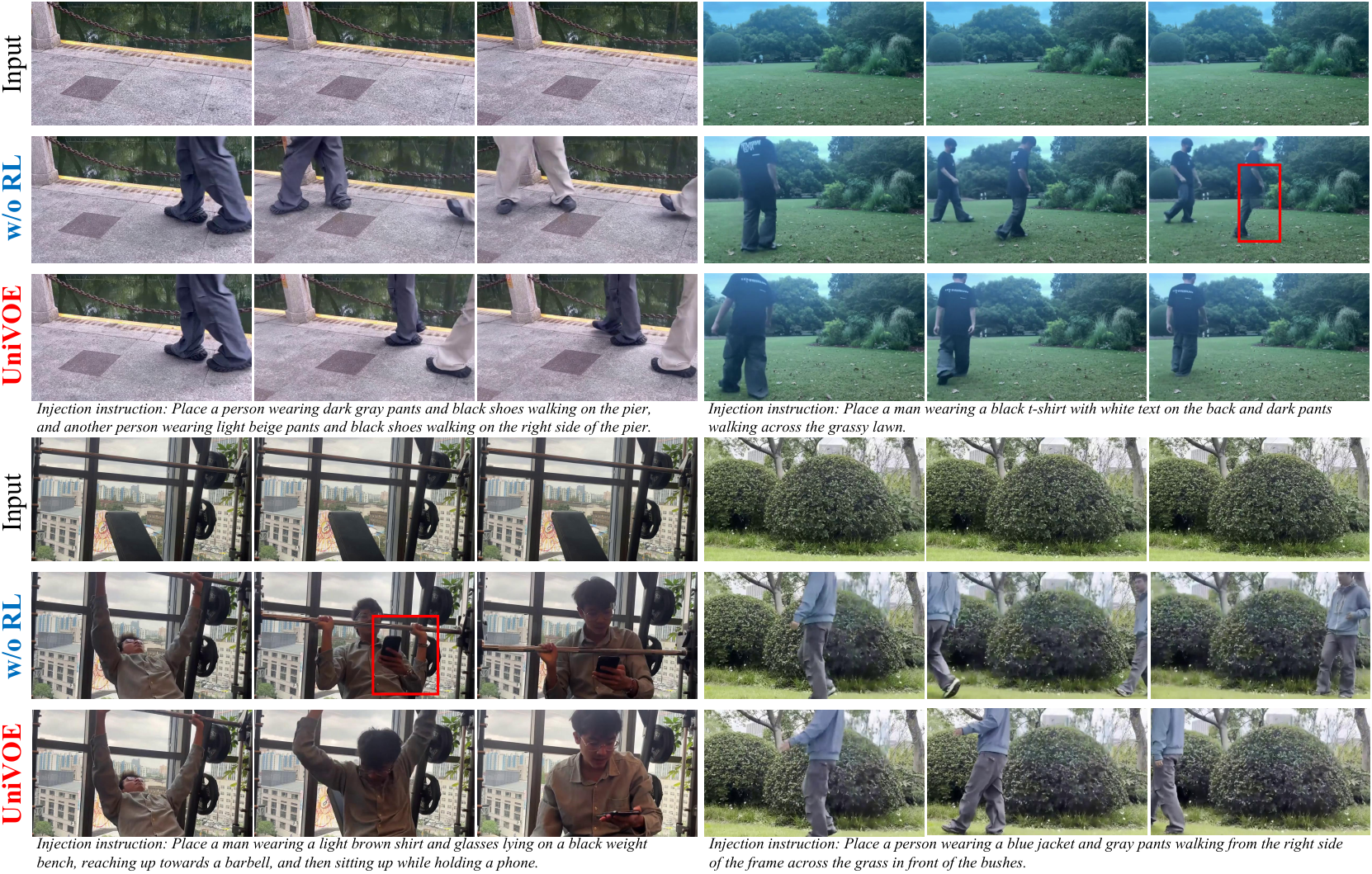}
  \vspace{-2.em}
  \caption{Ablation on the post-training stage of injector. We showcase 4 cases from input video and the edited results of ``w/o RL'' and ``UniVOE'', respectively. Related prompts are given below. Results of ``w/o RL'' exhibit color mutation (top-left), subject disappear (top-right), subject distortion (bottom-left), and semantic inconsistency (bottom-right). These issues can be well addressed in common scenarios by post-training.}
  \label{supp:fig:abla_inj}
  \vspace{-1.em}
\end{figure*}

\section{Ablation on Post-training Stage}
To validate the positive impact of post-training for video object injection, we compare performance between ``UniVOE'' and the ablation setting ``w/o RL'', the injector that is only pre-trained. In the main paper, we have provided quantitative results and mentioned that visual comparison is given in supplementary, here we provide qualitative results. We showcase visual comparison in Fig.~\ref{supp:fig:abla_inj}, ``w/o RL'' shows common issues exhibiting in video generation models, such as color mutation and unrealistic distortion, \textit{etc}. In most common scenes, these perceptually implausible cases can be well resolved by post-training as shown in Fig.~\ref{supp:fig:abla_inj}. This paper proves the crucial effectiveness of reward-based post-training for physical-realism of video object injection.

\section{More Results}
\begin{figure*}[t]
  \centering
    \includegraphics[width=1. \linewidth]{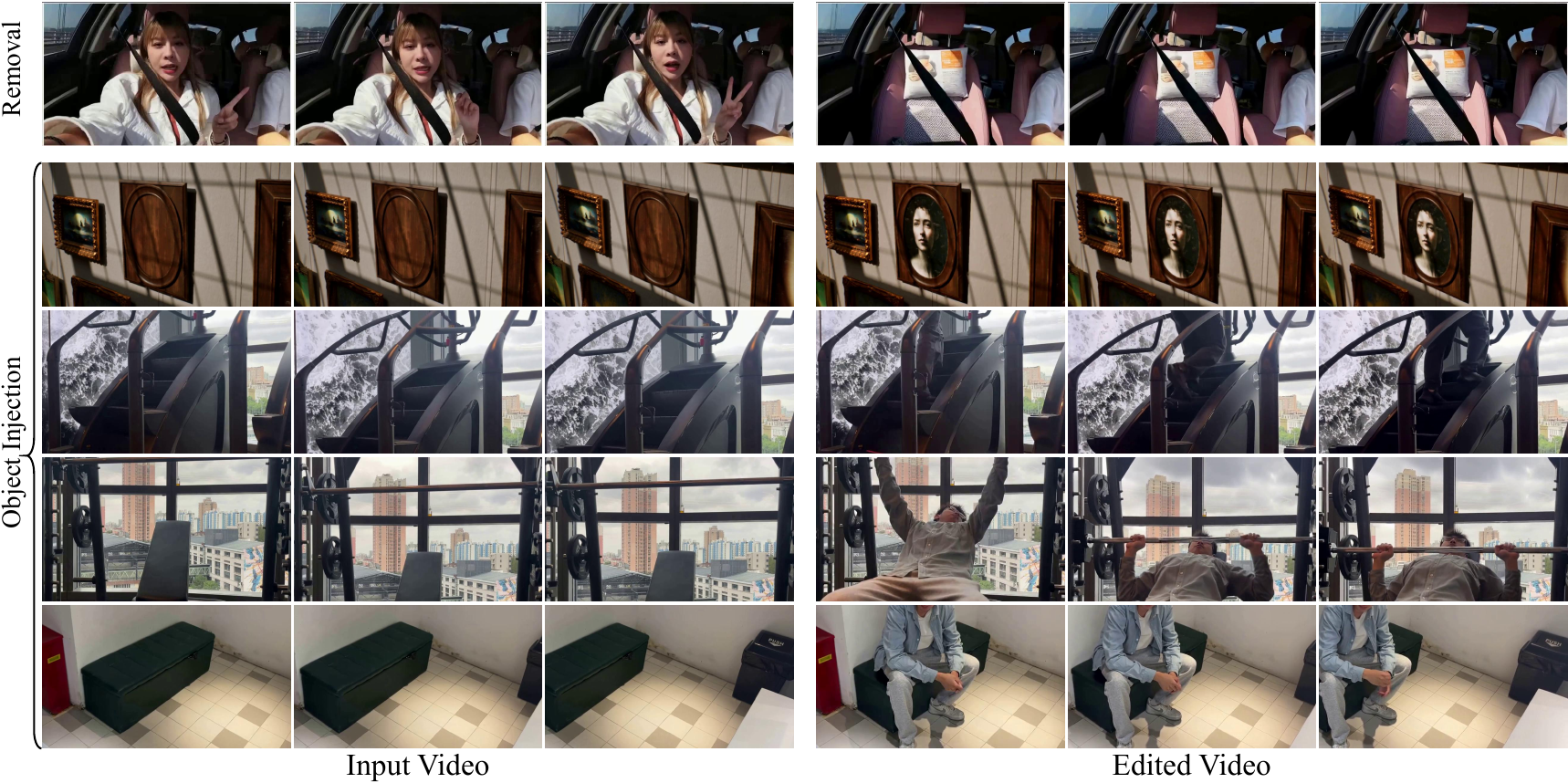}
  \vspace{-2.em}
  \caption{More results of video object removal and injection of UniVOE. For removal, UniVOE can fill realistic background content for the removed region. For injection, UniVOE not only inserts desired objects, but also predicts its natural interaction effects with background. Source videos can be found in supplementary materials.}
  \label{supp:fig:VOE_demo}
  \vspace{-1.5em}
\end{figure*}

In Fig.~\ref{supp:fig:VOE_demo}, we additionally provide some results of UniVOE, including object removal and injection. This further showcase superior performance and impressive generalization ability of UniVOE. The corresponding video demos can be found in our media supplement.

\section{Limitations and Future Work}
\label{supp:sec:discussion}

\paragraph{Dependence on the removal-derived bootstrap.}
The injector inherits whatever biases survive the filtering gate. The gate is strict-by-default and removes pairs with residual objects or corrupted background, but it cannot manufacture supervision that the remover never produces. Consequently, injection quality is upper-bounded by the coverage of the remover and the retained pairs. Designing more efficient post-training algorithms for injectors is an effective way to break through the performance ceiling imposed by the training set. This has been proven in our paper. In addition, jointly improving the remover and injector, or recycling high-reward injection rollouts back into the training pool, is also a natural way to break this dependence. We leave this cycle-like coordinated training strategy as a future work.

\paragraph{Reward-bounded post-training.}
GRAR aligns the injector with the preferences encoded in $Q_\theta$, so its ceiling is the fidelity of that reward model. Because the signed advantage in Eq.~(\ref{supp:eq:adv}) depends only on the within-group ranking rather than absolute scores, GRAR is robust to the reward scale and to occasional mis-scored rollouts. However, developing more diverse and preference-aligned reward model will further improve the quality of reward-guided post-training. Our team will continue to develop reward models for videos and apply and open-source them in future work.

\paragraph{Scope.}
Finally, our study targets object removal and injection on a single $23$B video backbone at $97$-frame $480$p. The removal-driven recipe and the GRAR post-training loss are not specific to injection, and applying them to these broader settings is left to future work.

\paragraph{On the lasting value of specialist editing models.}
One might argue that, as video foundation models keep scaling, future backbones will natively support universal human and object removal and injection, which would seem to diminish the value of dedicated specialist models such as ours. We take the opposite view, for two reasons that our results directly support. First, scale alone does not close the gap on vertical editing tasks: despite orders-of-magnitude fewer parameters, our $23$B specialist outperforms much larger general-purpose systems in our comparisons (e.g., the $200$B-parameter Seedance~2.0), indicating that task-specific data curation and post-training remain decisive for editing quality that generic base models do not yet deliver. Second, and more fundamentally, training a foundation model to edit natively requires millions of paired before/after editing supervision clips, which is infeasible to obtain by manual annotation alone. Industrial pipelines therefore rely on dedicated data operators that synthesize such pairs automatically. UniVOE is precisely such an operator: coupled with the strict quality gate, it can emit large volumes of high-quality object-present/object-absent and injection pairs, which are exactly the supervision a foundation model needs during its SFT stage. In this sense, specialist editing models are not superseded by stronger foundation models; they become the engines that supply the paired data on which those foundation models are trained. This reframes the contribution of UniVOE from a standalone editor to a reusable component of the broader video-model data ecosystem.

\end{document}